\pdfoutput=1

\documentclass[11pt]{article}

\usepackage{emnlp2021}

\usepackage{times}
\usepackage{latexsym}

\usepackage{graphics}
\usepackage{graphicx}
\usepackage{float}
\restylefloat{table}
\usepackage[T1]{fontenc}

\usepackage{amsfonts}

\usepackage[utf8]{inputenc}

\usepackage{microtype}

\title{KnowMAN: Weakly Supervised Multinomial Adversarial Networks}

\author{Luisa M\"{a}rz $^{\diamond,\dag}$, Ehsaneddin Asgari $^{\dag}$, Fabienne Braune $^{\dag}$,\\ \textbf{Franziska Zimmermann$^{\dag}$ and Benjamin Roth $^\diamond$}\\
$^\diamond$ Digital Philology, Research Group Data Mining and Machine Learning,\\ University of Vienna, Austria \\
$^\dag$ NLP Expert Center, Data:Lab, Volkswagen AG, Munich, Germany}

\begin{document}
\maketitle
\begin{abstract}
The absence of labeled data for training neural models is often addressed by leveraging knowledge about the specific task, resulting in heuristic but noisy labels.
The knowledge is captured in labeling functions, which detect certain regularities or patterns in the training samples and annotate corresponding labels for training. 
This process of weakly supervised training may result in an over-reliance on the signals captured by the labeling functions and hinder models to exploit other signals or to generalize well.
We propose KnowMAN, an adversarial scheme that enables to control influence of signals associated with specific labeling functions.
KnowMAN forces the network to learn representations that are invariant to those signals and to pick up other signals that are more generally associated with an output label.
KnowMAN strongly improves results compared to direct weakly supervised learning with a pre-trained transformer language model and a feature-based baseline.
\end{abstract}

\section{Introduction}
Neural approaches rely on labeled data sets for training.
For many tasks and languages, such data is either scarce or not available at all. 
Knowledge-based weak supervision tackles this problem by employing \emph{labeling functions (LFs)}. LFs are manually specified properties, e.g. keywords, that trigger the automatic annotation of a specific label.
However, these annotations contain noise and biases that need to be handled.

A recent approach for denoising weakly supervised data is Snorkel \citep{snorkel}.
Snorkel focuses on estimating the reliability of LFs and of the resulting heuristic \emph{labels}.
However, Snorkel does not address biases on the \emph{input side} of weakly supervised data, which might lead to learned representations that overfit the characteristics of specific LFs, hindering generalization.
We address the problem of overfitting to the LFs in this paper. 

Other approaches tackle such overfitting by deleting the LF signal completely from the input side of an annotated sample:
For example, \citet{go2009twitter} strip out emoticons that were used for labeling the sentiment in tweets, and \citet{alt2019improving} mask the entities used for distant supervision of relation extraction training data \cite{mintz2009distant}. 
However, as LFs are often constructed from the most prototypical and reliable signals (e.g., keywords), deleting them entirely from the feature space might -- while preventing over-reliance on them -- hurt prediction quality considerably.
However, we find a way to blur the signals of the LFs instead of removing them. 

In this work we propose KnowMAN (Knowledge-based Weakly Supervised Multinomial Adversarial Networks), a method for controllable \emph{soft deletion} of LF signals, allowing a trade-off between reliance and generalization.
Inspired by adversarial learning for domain adaptation \cite{MAN,ganin2015unsupervised}, we consider LFs as domains and aim to learn a \textit{LF-invariant} feature extractor in our model.
KnowMAN is composed of three modules: a feature extractor, a classifier, and a discriminator.
Specifically, KnowMAN employs a classifier that learns the actual task and an adversarial opponent, the LF- discriminator, that learns to distinguish between the different LFs.
Upstream of both is the shared feature extractor to which the gradient of the classifier and the reversed gradient of the discriminator are propagated.
In our experiments, the feature extractor for encoding the input is a multi-layer perceptron on top of either a bag-of-words vector or a transformer architecture, but KnowMAN is in principle usable with any differentiable feature extractor. 

KnowMAN consistently outperforms our baselines by 2 to 30\% depending on the dataset. By setting a hyperparameter $\lambda$ that controls the influence of the adversarial part we can control the degree of discarding the information of LF-specific signals. The optimal $\lambda$ value depends on the dataset and its properties.

The contributions of this work are
i) proposing an adversarial architecture for controlling the influence of signals associated with specific LFs,
ii) consistent improvements over weakly supervised baselines, 
iii) release of our code \footnote{\url{https://github.com/LuisaMaerz/KnowMAN}}.  
To our knowledge, we are the first that apply adversarial learning to overcome the noisiness of labels in weak supervision.

\section{Method}

\begin{figure}[]
    \centering
    \includegraphics[width=200pt]{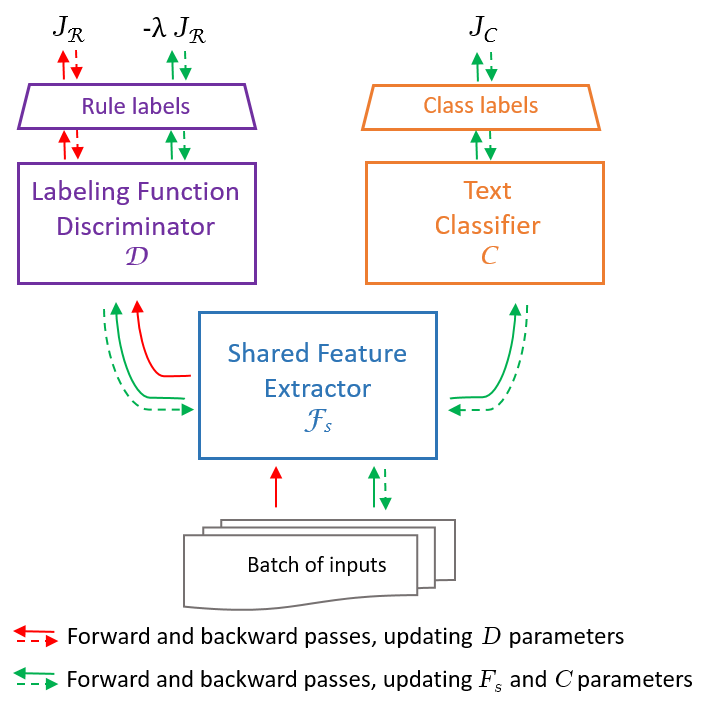}
    \caption{KnowMAN architecture. The figure depicts one iteration over a batch of inputs. The parameters of $\mathcal{C}$ and $\mathcal{F}_s$ are updated together, following the green arrows. The LF discriminator $\mathcal{D}$ is updated following the red arrows. Solid lines indicate forward, dashed lines the backward pass.}
    \label{model}
\end{figure}

Our approach is composed of three interacting modules i) the shared feature extractor $\mathcal{F}_s$, ii) the classifier $\mathcal{C}$ and iii) the LF discriminator $\mathcal{D}$. 
The loss function of $\mathcal{C}$ rewards the classifier $\mathcal{C}$ for predicting the correct label for the instance, and the gradient is used for optimizing the shared feature extractor and classifier modules towards that goal.
At the same time, the loss function for the LF-discriminator $\mathcal{D}$ rewards predicting which LF was responsible for labeling an instance. 
However, in adversarial optimization, KnowMAN backpropagates the \emph{reversed} gradient for the LF-discriminator, hence the information indicative for distinguishing between specific LFs is weakened throughout the network. 
The hyperparameter $\lambda$ is used to control the level of weakening the signals - the higher we choose the value the more influence is assigned to the discriminator information that goes into $\mathcal{D}$. 
The result of the interplay between classifier and LF-discriminator is a shared feature representation that is good at predicting the labels while reducing the influence of LF-specific signals, encouraging the shared feature extractor to take other information (correlated with all LFs for a class) into account. 

In Figure \ref{model}, the arrows illustrate the training
flow of the three modules. Due to the adversarial nature of the LF discriminator $\mathcal{D}$, it has to be trained with a separate optimizer (red arrows), while the rest of the network is updated with the main optimizer (green arrows). 
When $\mathcal{D}$ is trained the parameters of $\mathcal{C}$ and $\mathcal{F}_s$ are frozen and vice versa. 

To calculate the losses we utilize canonical negative log-likelihood loss (NLL) and use it for both, the classifier and the LF discriminator. 
The classification NLL can be formalized as: 
\begin{equation}
    \mathcal{L}_C(\hat{y_i}, y_i) = - \log P (\hat{y_i} = y_i)
\end{equation}
where $y_i$ is the (weakly supervised) annotated label and $\hat{y_i}$ is the prediction of the classifier module $\mathcal{C}$, for a training sample $i$.
Analogously, we can define the NLL for the LF discriminator: 
\begin{equation}
    \mathcal{L_D}(\hat{lf}_i, lf_i) = - \log P (\hat{lf}_i = lf_i)
\end{equation}
where $lf_i$ is the actual LF used for annotating sample $i$ and $\hat{lf}_i$ is the predicted LF by the discriminator $\mathcal{D}$. 
Accordingly, we minimize two different objectives within KnowMAN:
\begin{equation}
J_\mathcal{C} = \sum_{i=1}^{N} \mathcal{L_C} (\mathcal{C}(\mathcal{F}_s(x_i);y_i))
\end{equation}
\begin{equation}
J_\mathcal{D} = \sum_{i=1}^{N} \mathcal{L_D} (\mathcal{D}(\mathcal{F}_s(x_i);lf_i))
\end{equation}
Here the shared feature extractor has two different objectives: i) help $\mathcal{C}$ to achieve better classification performance and ii) make the feature distribution invariant to the signals from the LFs. This is captured by the shared objective:
\begin{equation}
J_{\mathcal{F}_s} = J_{\mathcal{C}} + \lambda \cdot (-J_{\mathcal{D}})
\end{equation}
where $\lambda$ is the parameter that controls the adversarial influence i.e. the degree of LF signal blur. $-J_{\mathcal{D}}$ is the reversed loss of the LF discriminator $\mathcal{D}$  that represents $\mathcal{C}s$ adversarial opponent.
In general, the exact implementation or architecture of the individual modules is interchangeable and can be set up as required. This makes KnowMAN a universally applicable and easily customizable architecture.

\section{Experiments}
\subsection{Data}
For our experiments we use three standard datasets for weak supervision.

\textbf{Spam.}
Based on the YouTube comments dataset \cite{spam} there is a smaller Spam dataset from Snorkel \cite{snorkel} where the task is to classify if a text is relevant to a certain YouTube video or contains spam. 
This dataset is very small and does consist of a train and a test set only. 
For the $10$ LFs keywords and regular expressions are used.

\textbf{Spouse.}
This dataset for extracting the \emph{spouse} relation has also been created by Snorkel, it is based on the Signal Media One-Million News Articles Dataset \cite{Corney2016WhatDA}. The $9$ LFs use information from a knowledge base, keywords and patterns. One peculiarity of this dataset is that over 90\% of the instances do not hold a spouse relation. 

\textbf{IMDb.} 
The IMDb dataset contains movie reviews that should be classified in terms of their sentiment (binary, positive or negative sentiment). The LFs used for this dataset are occurrences of positive and negative keywords from \cite{hu}.
A particular characteristic of this data set is the large amount of $6800$ LFs, which constitutes a particular challenge to the Snorkel denoising framework. 
As a result Snorkel fails to calculate its generative model, since its memory consumption exceeds the available limit of 32GB RAM.

\subsection{Experimental setup}
For the experiments we use two different methods for encoding the input: i) TF-IDF encoding and ii) a DistilBERT transformer.
For TF-IDF encoding, we vectorize\footnote{\url{https://scikit-learn.org/stable/modules/generated/sklearn.feature_extraction.text.TfidfVectorizer.html}} the input sentences and feed them to a simple MLP. 
In the transformer setting, the sequences of words are encoded using a pretrained DistilBERT. Similar to BERT \cite{devlin-etal-2019-bert}, DistilBERT is a masked transformer language model, which is a smaller, lighter, and faster version leveraging knowledge distillation while retaining 97\% of BERT's language understanding capabilities \cite{DistilBERT}. 

Our encoder takes the representation of the CLS token from a frozen DistilBERT and learns a non-linear transformation with a drop-out layer to avoid overfitting \cite{srivastava2014dropout}:
\[
h_i = DistilBERT(Sentence_i)_{[CLS]}
\]
\[
{F_s}_i = Dropout(ReLU(f(h_i)))
\]
where $DistilBERT(.)_{[CLS]}$ generates the hidden state of the BERT's classifier token (CLS) and the function $f$ represents a linear transformation for the $i^{th}$ sentence. 

The classifier and discriminator networks following the feature extractor are in line with the implementation of \citet{MAN} for domain-adversarial learning. Both are simple sequential models with dropout, batch normalization, $ReLU$ activation and softmax as the last layer. 
Please see our code for implementation details.
In the TF-IDF setup we use Adam \cite{adam} for both optimizers. When using transformer encoding the $\mathcal{D}$ optimizer again is Adam and the $\mathcal{C}$ optimizer is AdamW \cite{adamW}, as this yielded more stable results.

\textbf{Baselines}
For each input encoding we implemented several baselines. 
Weakly supervised TF-IDF (\textit{WS TF-IDF}) and Weakly supervised DistilBERT (\textit{WS DistilBERT}). Both calculate the labels for each instance in the train set based on their matching LFs. \textit{WS TF-IDF} directly applies a logistic regression classifier to the input and the calculated labels. \textit{WS DistilBERT} directly uses the DistilBERT uncased model for English \cite{DistilBERT} as a prediction model.
The second baseline (\textit{Feature TF-IDF}, \textit{Feature DistilBERT}) uses feature extractor and classifier layers of KnowMAN  without taking the information of $\mathcal{D}$ into account (this is equal to setting $\lambda$ to zero). 
We also fine-tuned the pure language model (\textit{Fine-tuned DistilBERT}) without further transformations and without integrating the KnowMAN architecture.

We also compare with training \textit{TF-IDF} and \textit{DistilBERT} models on labels denoised by Snorke (\textit{Snorkel TF-IDF}, \textit{Snorkel DistilBERT}). 
However, Snorkel denoising failed for the IMDb data set due to the large amount of LFs.

\begin{table*}[!htbp]
    \centering
    \begin{tabular}{l|c|ccc|c}
         & Spam & &  Spouse &  & IMDb  \\ \hline
         & Acc & P & R & F1 & Acc  \\ \hline
      WS TF-IDF   & 0.87 &  0.12 & \textbf{0.83} & 0.20* & 0.65* \\
      Feature TF-IDF & 0.91 & 0.12 & 0.76 & 0.21* & 0.75* \\ 
      Snorkel TF-IDF & 0.81 & 0.18 & 0.63 & 0.28* & 0.50*  \\ 
      KnowMAN TF-IDF & \textbf{0.94} & \textbf{0.16} & 0.72 & \textbf{0.35} & \textbf{0.77} \\ \hline 
      Fine-tuned DistilBERT & \textbf{0.92} & 0.14 & 0.78 & 0.24 & 0.70 \\
      WS DistilBERT & 0.87 & 0.09 & \textbf{0.90} & 0.17* & 0.67* \\ 
      Feature DistilBERT & 0.86 & 0.18 & 0.80 & 0.29* & 0.74 \\ 
      Snorkel DistilBERT & 0.88 & 0.13 & 0.70 & 0.23* & 0.49* \\
      KnowMAN DistilBERT & 0.90 & \textbf{0.27} & 0.67 & \textbf{0.39} & \textbf{0.76} \\ \hline \hline 
    \end{tabular}
    \caption{Results on the test sets. The * indicates that KnowMAN performs significantly better than the marked model. For the Spouse data set we do report significance for the F1 scores only.}
    \label{res}
\end{table*}

\textbf{KnowMAN}
We refer to the KnowMAN architecture as \textit{TF-IDF KnowMAN} and \textit{DistilBERT KnowMAN}.
Depending on the dataset we choose different $\lambda$ values. 
We also implemented two ways of evaluation and best model saving during training: i) evaluate after each batch and save the best model, ii) evaluate after a certain number of steps in between the batches and save the best model.

\textbf{Hyperparameters}
We perform hyperparameter tuning using Bayesian optimization \citep{snoek2012} for the IMDb and Spouse datasets. For Spam, hyperparameters are not optimized, as no validation set is available. Sampling history and resulting hyperparameters are reported in the Appendix, Figures \ref{fig:imdb_hyperparam}, \ref{fig:spouse_hyperparam} as well as hyperparameters chosen for the Spam data set. 

\textbf{Evaluation}
For the evaluation of the IMDb and the Spam datasets we use accuracy, for the Spouse dataset we use the macro F1 score of the positive class.
To check statistical significance we use randomized testing \cite{yeh-2000-accurate}. Results are considered significant if $\rho$ < 0.05.

\subsection{Results}

The results of the experiments are shown in Table \ref{res}.
For the TF-IDF setup \textit{KnowMAN TF-IDF} outperforms the baselines across all datasets. We find the optimal $\lambda$ values as follows: Spam/Spouse/IMDb = 2/5/4.9.  
Using the additional feature extractor layer (\textit{Feature TF-IDF}) is beneficial compared to direct logistic regression for all datasets.
\textit{Snorkel TF-IDF} can outperform the other two baselines for the Spouse dataset only. 

Fine tuning of DistilBERT can not outperform our best KnowMAN. However, for the Spam dataset \textit{Fine-tuned DistilBERT} gives better results than \textit{KnowMAN DistilBERT} but still is worse than \textit{KnowMAN TF-IDF}.
Using \textit{WS DistilBERT} gives the same results for the Spam dataset and slightly better results for IMDb, when compared to \textit{WS TF-IDF}, for Spouse the performance decreases. 
\textit{Snorkel DistilBERT} can outperform the other two baselines for the Spam dataset only.
The low performance of Snorkel on IMDb (for both DistilBERT and TF-IDF) might be explained by the very large amount of LF for this dataset.
The \textit{KnowMAN DistilBERT} results across datasets are in line with the TF-IDF setup - KnowMAN can outperform all baselines for the Spouse and IMDb dataset. We observe that $\lambda = 5$ for Spouse and $\lambda = 1$ for IMDb is most beneficial when using DistilBERT. For the Spam dataset we observe that KnowMAN (with $\lambda = 2 $) outperforms all the baselines, except for the fine-tuned DistilBERT model. 

\textbf{Discussion}
The performance drop we observe with DistilBERT for KnowMAN compared to the tf-idf setup of the IMDb dataset could be explained by implementation details. Due to memory issues we have to truncate the input when using DistilBERT. Since the movie reviews from IMDb are rather long this could harm performance. 
Since the Spam dataset is very small a single wrongly classified instance can have great impact on the results. This could explain why \textit{KnowMAN TF-IDF} outperforms \textit{KnowMAN DistilBERT} here as well. 
In general we could not perform hyperparameter optimization for the DistilBERT experiments due to memory issues. Therefore the results for that experiments might not have reached their optimum. However, the results show the value of using KnowMAN though.
Overall our results confirm the assumption that KnowMAN enables a focus shift of the shared feature extractor from the signals of the LFs towards signals of other valuable information. KnowMAN consistently improves over the other experiments significantly - except for the Spam dataset. We assume that the dataset size is too small to see significant changes in the results. 
Compared to the implementation of \citet{MAN} we could not use the specialized domain feature extractor for our datasets in the experiments. This is due to the fact that our test sets do not contain information about LF matches. However, we will address this issue by integrating a mixture of experts module for the specialized feature extractor as recommended by \citet{chen-etal-2019-multi}.

\section{Related Work}
Adversarial neural networks have been used to reduce the divergence between distributions, such as \citet{gan2014}, \citet{chen2018} and \citet{ganin2015unsupervised}. 
The latter proposed an architecture for gradient reversal and a shared feature extractor. Unlike us, they focused on a binary domain discriminator.
Similarly, \cite{MAN} use an adversarial approach in a multinomial scenario for domain adaptation.

Some works on adversarial learning in the context of weak supervision focus on different aspects and only share similarity in name with our approach:
\citet{wu2017adversarial} use \emph{virtual adversarial training} \cite{miyato2016adversarial} for perturbing input representations, which can be viewed as a general regularization technique not specific to weakly supervised learning.
\citet{qin2018dsgan, zeng2018adversarial} use generative adversarial mechanisms for selecting \emph{negative} training instances that are difficult to discriminate from heuristically annotated ones for a classifier.

Several approaches have focused on denoising the labels for weakly supervised learning \cite{takamatsu2012reducing,manning2014stanford,lin2016neural}.
Snorkel \cite{snorkel} is one of the most general approaches in this line of work.
However, Snorkel only models biases and correlations of LFs, and does not consider problems of weak supervision that may stem from biases in the features and learned representations.

A recent approach that focuses on denoising weakly supervised data is \cite{knodle}. Knodle is a framework for comparison of different methods that improve weakly supervised learning. We use some of their datasets for our approach but denoise the signals of the LFs during training.

\section{Conclusion}
We propose KnowMAN - an adversarial neural network for training models with noisy weakly supervised data. 
By integrating a shared feature extractor that learns labeling function invariant features, KnowMAN can improve results on weakly supervised data drastically across all experiments and datasets in our setup.
The experiments also show that the adverse effect of labeling function-specific signals is highly dependent on the datasets and their properties. Therefore, it is crucial to fine-tune the $\lambda$ parameter on a validation set to find the optimal degree of blurring the labeling function signals. Since the modules in the KnowMAN architecture are easily exchangeable, KnowMAN can be applied to any architecture and dataset labeled with heuristic labeling functions.

\section*{Acknowledgements}
This research was funded by the WWTF through theproject ”Knowledge-infused Deep Learning for Nat-ural Language Processing” (WWTF Vienna ResearchGroup VRG19-008), by the Deutsche Forschungs-gemeinschaft (DFG, German Research Foundation) -RO 5127/2-1.

\bibliography{anthology,custom}
\bibliographystyle{acl_natbib}

\appendix

\section{Appendix}
\label{sec:appendix}

\subsection{Dataset statistics}
The datasets used for the KnowMAN experiments have different properties. Especially the numer of labeling functions and the dataset sizes varies a lot.

\begin{table}[h!]
    \centering
    \begin{tabular}{c|c|c|c}
         \textbf{dataset} & \textbf{classes} & \textbf{train/test samples} & \textbf{lfs}  \\ \hline
         Spam & 2 & 1586/250 & 10 \\
         Spouse & 2 & 22254/2701 & 9 \\
         IMDb & 2 & 40000/5000 & 6786 \\ \hline
    \end{tabular}
    \caption{Dataset statistics for KnowMAN experiments. Lfs are labeling functions.}
    \label{tab:my_label}
\end{table}

\subsection{Hyperparameter optimization}

We perform hyperparameter tuning using Bayesian optimization \citep{snoek2012}. Bayesian Optimization is an approach that uses the Bayes Theorem to direct the search in order to find the minimum or maximum of a black-box objective function. In comparison with random search and grid search, it tends to obtain better hyperparameters in fewer steps by making a proper balance between exploration and exploitation steps.
Our hyperparameter space includes batch size, dropout, number of iterations over $\mathcal{D}$, the shared hidden size of the models, learning rate for $\mathcal{D}$ and $\mathcal{F}_s, \mathcal{C}$ and the number of layers of $\mathcal{C}, \mathcal{D}$ and $\mathcal{F}_s$.
We implemented two ways of evaluation and best model saving during training: i) evaluate after each batch and save the best model, ii) evaluate after a certain number of steps in between the batches and save the best model. We also optimized the number of steps if logging in between a batch. 

We evaluated the models for IMDb and Spouse on the respective validation set. For the Spam dataset, there is no development set available and we used the following hyperparameters for \textit{KnowMAN TF-IDF} following the parameters used in \citet{chen-cardie-2018-multinomial}: \textbf{Batch size:} 32, \textbf{dropout:} 0.4, \textbf{n critic:} 5, \textbf{lambda:} 2.0, \textbf{shared hidden size:} 700, \textbf{learning rate C \& F:} 0.0001, \textbf{learning rate D:} 0.0001 , \textbf{number of F layers:} 1, \textbf{number of C layers:} 1, \textbf{number of D layers:} 1.

\begin{figure*}[h!]
    \centering
    \includegraphics[width=450pt]{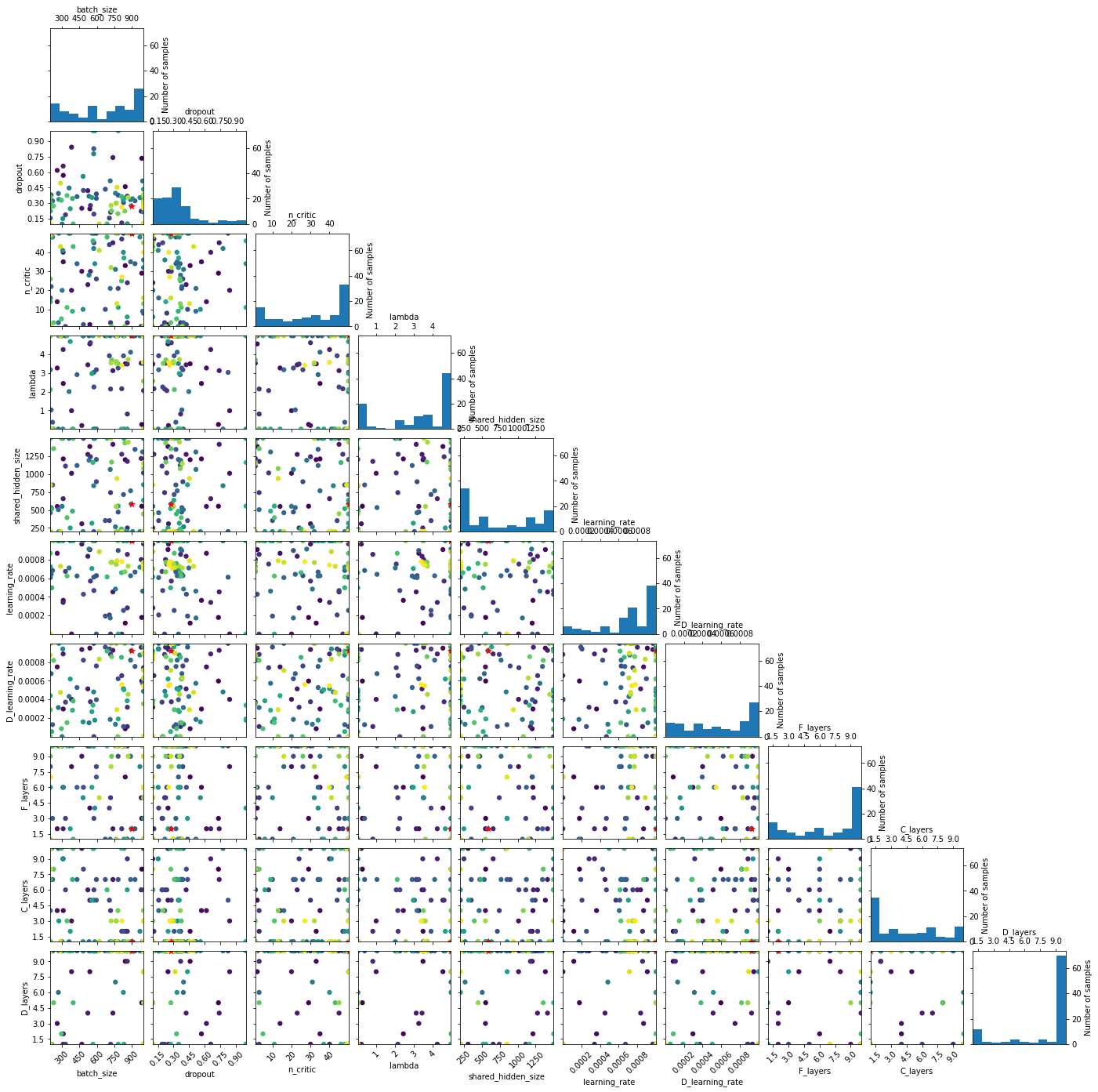}
    \caption{Sampled hyperparameters for KnowMAN TF-IDF on IMDb. Optimal hyperparameters are indicated in red.\\
    \textbf{Batch size:} 895, \textbf{dropout:} 0.275, \textbf{n critic:} 50, \textbf{lambda:} 4.9, \textbf{shared hidden size:} 585, \textbf{learning rate C \& F:} 0.0001, \textbf{learning rate D:} 0.0001, \textbf{number of F layers:} 1 , \textbf{number of C layers:} 1, \textbf{number of D layers:} 10. \\
    Histograms on the diagonal show how, for each hyperparameter, how many samples have been drawn during optimization.}
    \label{fig:imdb_hyperparam}
\end{figure*}

\begin{figure*}
    \centering
    \includegraphics[width=450pt]{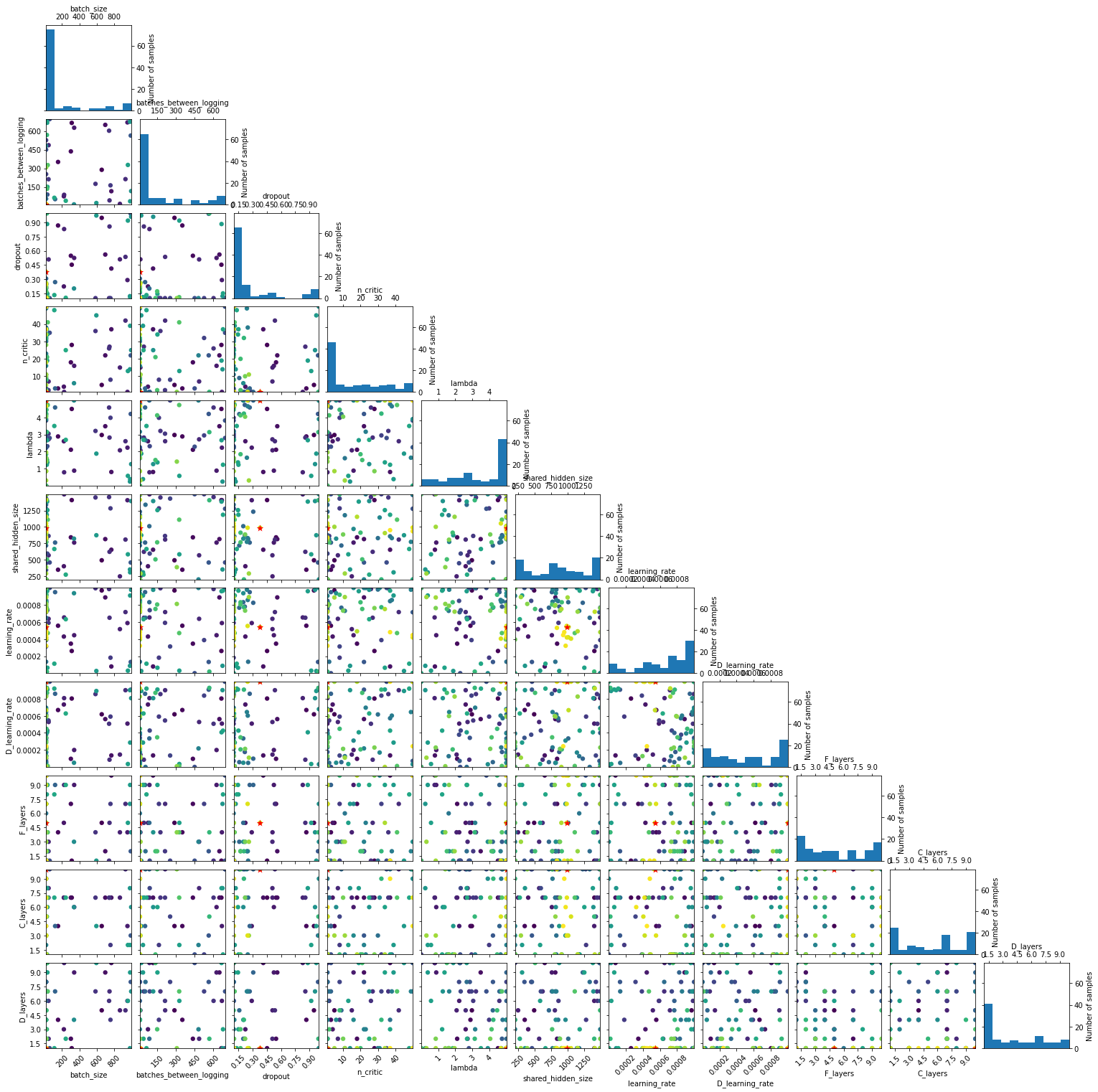}
    \caption{Sampled hyperparameters for KnowMAN DistilBERT on Spouse. Optimal hyperparameters are indicated in red.\\
    \textbf{Batch size:} 16, \textbf{dropout:} 0.379, \textbf{n critic:} 1, \textbf{lambda:} 5.0, \textbf{shared hidden size:} 988, \textbf{learning rate C \& F:} 0.0005, \textbf{learning rate D:} 0.001 , \textbf{number of F layers:} 5, \textbf{number of C layers:} 10, \textbf{number of D layers:} 1. \\
    Histograms on the diagonal show how, for each hyperparameter, how many samples have been drawn during optimization.}
    \label{fig:spouse_hyperparam}
\end{figure*}

\subsection{Experimental details}
We ran our experiments on a DGX-1 server with one V100 GPU per experiment. The runtime of one model depends on the dataset:  0.25 hours for the Spam dataset, 0.25 hours for the Spouse dataset, and 8 hours for the IMDb dataset.

Please find our implementation at \url{https://github.com/LuisaMaerz/KnowMAN}.

\end{document}